\documentclass[11pt]{article}

\usepackage[final]{acl}

\usepackage{times}
\usepackage{latexsym}
\usepackage[table]{xcolor}

\usepackage[T1]{fontenc}

\usepackage[utf8]{inputenc}
\usepackage{amsmath}
\usepackage{subcaption}
\usepackage{booktabs}
\usepackage{amsthm}

\usepackage{microtype}
\usepackage{cleveref}
\usepackage{multirow}

\usepackage{inconsolata}

\usepackage{graphicx}

\title{Distribution Shift Alignment Helps LLMs Simulate Survey Response Distributions}

\author{
  \textbf{Ji Huang\textsuperscript{\(\spadesuit\)}\thanks{Equal contribution.}},
  \textbf{Mengfei Li\textsuperscript{\(\heartsuit\)}\footnotemark[1]},
  \textbf{Shuai Shao\textsuperscript{\(\spadesuit\)}\thanks{Corresponding author.}}
\\
  \textsuperscript{\(\spadesuit\)}School of Computer Science, University of Science and Technology of China \\
  \textsuperscript{\(\heartsuit\)}School of Management, Fudan University
\\
    \texttt{huangjustc@mail.ustc.edu.cn}\hspace{0.9em}
\texttt{mfli22@m.fudan.edu.cn}\hspace{0.9em}
\texttt{shao10@ustc.edu.cn}
}

\begin{document}
\maketitle
\begin{abstract}
Large language models (LLMs) offer a promising way to simulate human survey responses, potentially reducing the cost of large-scale data collection. However, existing zero-shot methods suffer from prompt sensitivity and low accuracy, while conventional fine-tuning approaches mostly fit the training set distributions and struggle to produce results more accurate than the training set itself, which deviates from the original goal of using LLMs to simulate survey responses. Building on this observation, we introduce Distribution Shift Alignment (DSA), a two-stage fine-tuning method that aligns both the output distributions and the distribution shifts across different backgrounds. By learning how these distributions change rather than fitting training data, DSA can provide results substantially closer to the true distribution than the training data. Empirically, DSA consistently outperforms other methods on five public survey datasets. We further conduct a comprehensive comparison covering accuracy, robustness, and data savings. DSA reduces the required real data by 53.48-69.12\%, demonstrating its effectiveness and efficiency in survey simulation.
\end{abstract}

\section{Introduction}

\begin{figure}[t]
\centering
  \includegraphics[width=0.98\linewidth]{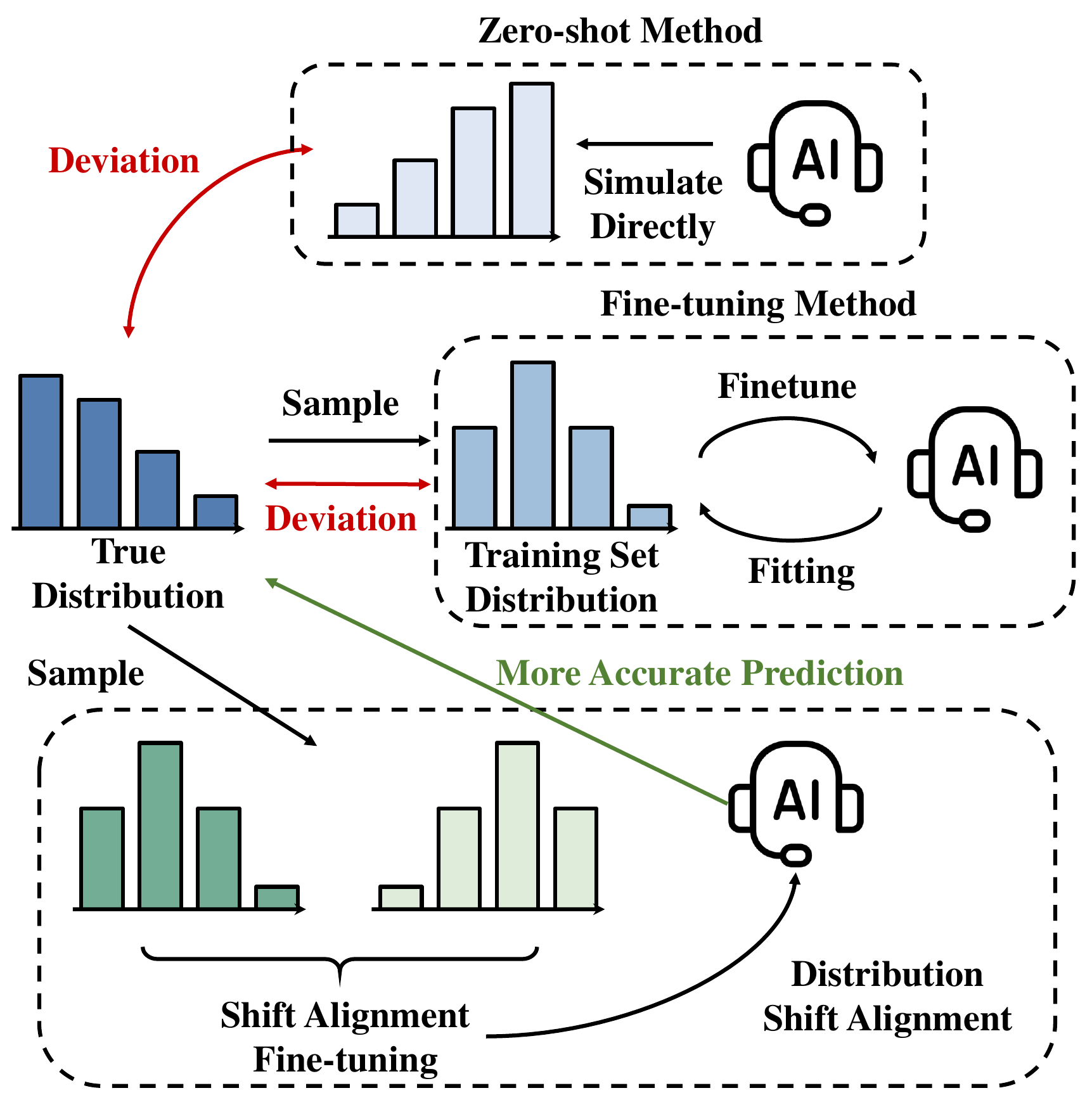}
  \caption {Overview of different approaches for simulating survey responses.}
    \label{fig:intro}
\end{figure}

Surveys are essential tools in social sciences, market research, and political science, providing valuable data for decision-making. However, large-scale surveys are costly and resource-intensive. Large language models (LLMs), trained on vast amounts of human text, offer a promising alternative by simulating human responses in surveys. This has led to growing research on whether LLMs can accurately replicate human preferences, potentially reducing the costs of data collection. However, existing methods largely focus on fitting the training data, and fail to outperform the empirical training distribution. In this work, we study how to leverage LLMs to move beyond this ceiling.

One existing approach is zero-shot methods, where LLMs generate survey responses without fine-tuning on real data \cite{zeroshot}, as shown in \Cref{fig:intro}. Further research includes using prompt engineering to better simulate \cite{PE,PE1}. While zero-shot methods are data-efficient, their results often deviate from real-world distributions. Additionally, these methods are sensitive to prompt changes, leading to varying results \cite{nature}.

Another approach is fine-tuning methods. AAE \cite{AAE} proposed fine-tuning a conditional probability model to align the LLM's output distribution with the training set. Two recent studies \cite{TKFT1,TKFT2} directly fine-tuned the LLM's token probabilities to match the training set's distribution. While fine-tuning produces more accurate results than zero-shot methods, its improvement mainly comes from aligning the LLM with the training set distribution, and is limited to reproducing the patterns of the training data without achieving higher accuracy, as shown in \Cref{fig:intro}. This raises the question of why not directly use the training set distribution as result, as these methods underutilize the LLM's potential.

Considering these limitations, we explore how to leverage LLMs' capabilities to yield results closer to the true distribution than training data. 
We find that, although LLMs may not be able to predict the exact distribution of human preferences, they are effective at identifying the differences in preferences across different backgrounds. For example, LLMs cannot accurately predict how many people like sports cars, but it can capture the trend that younger people prefer sports cars more than middle-aged people. This ability to predict preference shifts can align the distribution from different backgrounds, and integrate information from diverse backgrounds.

Building on this idea, we propose using the Distribution Shift Alignment (DSA) fine-tuning to help LLMs simulate human choice distributions. Our approach starts by collecting a small amount of training data and using it to fine-tune the LLM's token probabilities corresponding to survey options. The fine-tuning process is divided into two stages. In the first stage, we fine-tune the LLM directly using the training data, aligning its output distribution with that of the training data. This stage helps the LLM’s output better match the distribution of the training data and corrects biases. In the second stage, we train the LLM to align the distribution shifts between different backgrounds using a designed distribution shift loss. By fine-tuning the LLM to maintain the same differences between the distributions of different backgrounds, and using these differences to estimate the distribution of certain background from all other backgrounds, we can obtain a distribution closer to the real one than the training data, as shown in \Cref{fig:intro}. In the \Cref{sec:appendixtheory}, we prove that this estimation is unbiased with the true distribution under strong assumption.

In practice, we study two tasks: \emph{Enhancement} (data-efficient improvement for the same question) and \emph{Generation} (zero-adaptation transfer to unseen questions and/or populations). We evaluated DSA on five publicly available survey datasets in different areas, and compare DSA with both zero-shot and fine-tuning methods. DSA consistently outperformed all others in accurately simulating the real distribution. Using DSA, we were able to reduce the amount of data required by 53.48\%-69.12\% to achieve similar accuracy, significantly cutting down the costs. In general, this paper makes the following contributions:

\begin{itemize}
\item We propose Distribution Shift Alignment fine-tuning for simulating human survey responses with LLMs, which can generate results closer to the real distribution than the training data.

\item We incorporate the training-set distribution as a baseline to compare with fine-tuning methods, enabling evaluations of multiple characteristics—such as data savings—that were often ignored in previous studies.

\item We conducted experiments on five datasets under both the \emph{Enhancement} and \emph{Generation} tasks, showing that DSA provides the most accurate predictions and reduces the cost of conducting surveys by up to 53.48\%-69.12\%.

\end{itemize}

\section{Methodology}

\begin{figure*}[t]
\centering
  \includegraphics[width=0.95\linewidth]{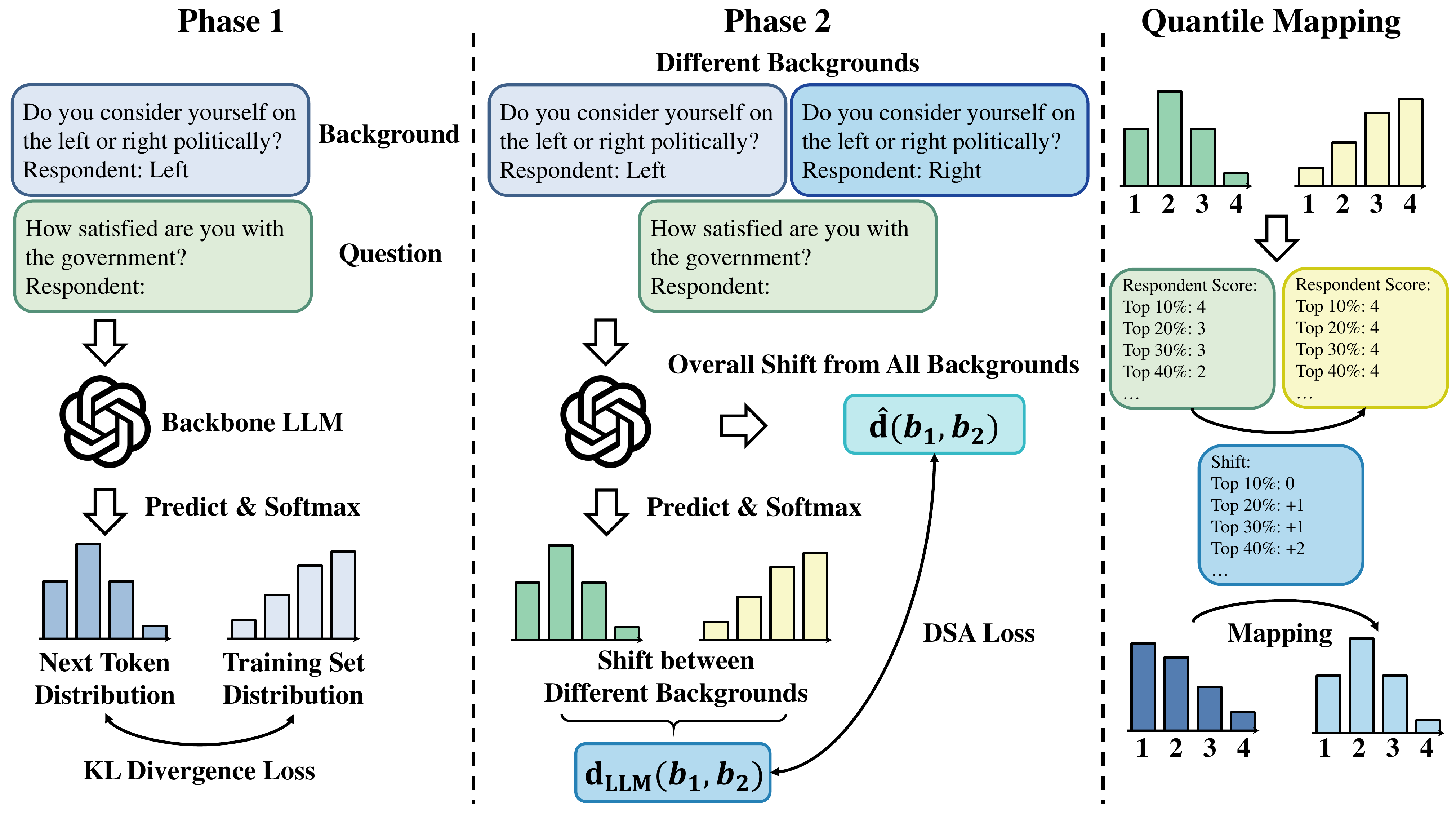}
  \caption {Framework of our proposed distribution shift alignment fine-tuning method.}
    \label{fig:framework}
\end{figure*}

\subsection{Task Definition}

In a survey, respondents’ choices on a core question $C$ form a probability distribution over its options
$P(C) = [p_1, p_2, \dots, p_n]$.
To model respondents' backgrounds, we include background questions
$B = (B_1, B_2, \dots, B_m)$, where each $B_i = (b_{i1}, b_{i2}, \dots, b_{iN_i})$.
Given a background attribute combination $\mathbf{b} \in B_1 \times \dots \times B_m$,
the goal is to predict the conditional choice distribution $P(C \mid \mathbf{b})$. We study two widely applicable tasks in real-world:

\paragraph{Enhancement.}
For a target question $C^{\star}$, we first collect a small set $\mathcal{D}_{\text{cal}}^{(k)}(C^{\star})$ and further fine-tune the LLM on it,
then evaluate on held-out data from the same question to estimate $\widehat{P}(C^{\star}\mid \mathbf{b})$.
This corresponds to scenarios where a new survey question has limited responses, and we want to improve estimation using a small amount of collected data.

\paragraph{Generation.}
We fine-tune the LLM on a training set $\mathcal{D}_{\text{train}}$ and directly test it on an unseen dataset $\mathcal{D}_{\text{test}}$
with different questions and/or different respondent backgrounds, requiring zero additional adaptation, to estimate $\widehat{P}(C\mid \mathbf{b})$.
This matches applications where one aims to transfer survey simulation to new questions or populations without collecting target-domain labels.

Prior work exclusively focuses on the Generation setting, while Enhancement is also a practically valuable task.

\subsection{Phase 1: Aligning with Training Set Distributions}

For both tasks, given the training data, we can compute the choice distribution of different backgrounds \(P^{\text{Train}} (C \mid \mathbf{b})\), which is a sample of the true distribution. In the first stage of fine-tuning, we focus on aligning the model’s outputs with the observed distributions from the training data. To achieve more precise fine-tuning, we adopt the same approach as the token-level fine-tuning \cite{TKFT1,TKFT2}, directly fine-tuning the LLM at the token level. Specifically, we treat the background information as prompt, and then assign a label to each option \(c_i\) of the core question (e.g., score from 1 to 5), asking the LLM to output the corresponding label directly, as shown in \Cref{fig:framework}. Then, the probability of the token \(t_i\) corresponding to the option label is regarded as the predicted probability:
\[
P_i^{\text{LLM}} = \frac{e^{t_i}}{\sum_{j=1}^{n} e^{t_j}},
\]
We train the LLM to align its output distribution with the observed distribution in the training data using the KL divergence loss:
\[
\mathcal{L} = \sum_{\mathbf{b} \in B_1 \times \dots \times B_m} P^{\text{LLM}}(C \mid \mathbf{b}) \log \left( \frac{P^{\text{LLM}}(C \mid \mathbf{b})}{P^{\text{Train}}(C \mid \mathbf{b})} \right)
\]

This phase of fine-tuning aligns the LLM’s output distribution with the training data distribution.

\subsection{Phase 2: Aligning Distribution Shifts across Backgrounds}

In the second stage of fine-tuning, we align the influence of each background question on the LLM output distributions. Specifically, assume that each background question is independent of the others. Then, when all other backgrounds are held constant, changing the choice on a particular background question (e.g., rural vs. urban) should cause the same shift in the distribution of the core question. By aligning these shifts in the LLM's output distribution, we can obtain a better estimate of the true distribution, as shown in \Cref{fig:framework}. Stage 1 fits each subgroup independently to its empirical distribution, which is noisy for rare subgroups. Stage 2 adds a cross-background consistency constraint by pooling single-attribute shifts across many subgroup pairs, providing a lower-variance supervision signal than directly fitting each small subgroup alone.

To mathematically model this shift, we use quantile mapping as shown in \Cref{fig:framework} to describe the difference between different distributions. Specifically, we rank respondents based on the scores they give to the core question and compute the scores given by respondents at different quantiles (e.g., the score given by the top 10\% of respondents in the \(\mathbf{b}_1\) background is denoted as \(s^{10\%}_{\mathbf{b}_1}\)). We then use the differences between different backgrounds of the same quantile to represent the shift between the two distributions:
\[
\mathbf{d}(\mathbf{b}_1, \mathbf{b}_2) = [d_k(\mathbf{b}_1, \mathbf{b}_2)] = [s^k_{\mathbf{b}_1} - s^k_{\mathbf{b}_2}]_{k = 0\%, 10\%,\ \dots}
\]

After Stage~1, we estimate the \emph{distribution shift} induced by a single background change.
For each pair of backgrounds $(\mathbf{b}_1,\mathbf{b}_2)$ with $\text{distance}(\mathbf{b}_1,\mathbf{b}_2)=1$, we compute a quantile-mapping operator
$T_{\mathbf{b}_1\rightarrow\mathbf{b}_2}$ from the Stage~1 predicted distributions, which maps a reference distribution under $\mathbf{b}_1$ to a target distribution under $\mathbf{b}_2$.
Importantly, this operator is estimated using \emph{all} available instances that match the corresponding single-attribute change, rather than relying on a specific pair, yielding a more stable shift estimate.

We then construct training pairs by randomly sampling two virtual respondents whose backgrounds differ in exactly one attribute, and obtain the predicted distributions of LLM
$P_{\text{LLM}}(C\mid \mathbf{b}_1)$ and $P_{\text{LLM}}(C\mid \mathbf{b}_2)$.
The quantile-mapping operator is used to form a \emph{target distribution}:
\[
\tilde{P}(C\mid \mathbf{b}_2) \;=\; T_{\mathbf{b}_1\rightarrow\mathbf{b}_2}\!\left(P_{\text{LLM}}(C\mid \mathbf{b}_1)\right).
\]
The Stage~2 objective aligns the LLM prediction under $\mathbf{b}_2$ to this target distribution:
\[
\mathcal{L}_{\text{DSA}}
=\sum_{(\mathbf{b}_1,\mathbf{b}_2)}
\mathcal{L}_{\mathrm{KL}}\!\left(
P_{\text{LLM}}(C\mid \mathbf{b}_2)\; \big\|\; \tilde{P}(C\mid \mathbf{b}_2)
\right).
\]

Because quantile mapping is non-differentiable, we do not directly supervise the (signed) shift vector. 
Instead, we construct a target distribution $\tilde{P}(C\mid\mathbf{b}_2)$ by applying the mapping to $P_{\text{LLM}}(C\mid\mathbf{b}_1)$, and align $P_{\text{LLM}}(C\mid\mathbf{b}_2)$ to this target; gradients are backpropagated only through the smaller subgroup prediction. This concentrates updates on data-scarce, high-variance subgroups: large subgroups already yield stable low-variance estimates, while small subgroups benefit most from stronger learning signals. We use quantile mapping because most survey options are ordinal. It provides an interpretable shift representation along the ordered response axis and remains stable under sparse subgroup data. In highly imbalanced cases, it degenerates to a near-identity mapping instead of introducing arbitrary shifts.

We implement fine-tuning by updating only the last transformer block and the output softmax layer, which is sufficient for distribution calibration while reducing overfitting and training cost.

\subsection{Comparison with Prior Work}

Our work is closely related to two recent studies of token fine-tuning (TKFT), which fine-tune LLMs' distribution on token-level to simulate survey distribution \cite{TKFT1,TKFT2}, but differs in task definition and target granularity. 
Those works focus on the task of \textbf{Generation}: training on many question--population pairs and evaluating on unseen questions and/or populations without further adaptation. 
We additionally study \textbf{Enhancement}: given a target question $C^\star$ and a small set of real survey data, the model is further adapted to improve estimation on the \emph{same} question. 
However, in the enhancement setting, TKFT often shows limited gains over the training data, which can reduce its practical impact in real-world use.

\begin{table*}[]
\centering
\resizebox{.99\linewidth}{!}{
\begin{tabular}{l|l|cccccc|cccccc}
\toprule
                            &                             & \multicolumn{6}{c|}{WD$\downarrow$}                                                                                                                                                                                                                                                  & \multicolumn{6}{c}{JSD$\downarrow$}                                                                                                                                                                                                                                                  \\ \cmidrule(lr){3-14}
\multirow{-2}{*}{LLM}       & \multirow{-2}{*}{Method}    & ESS11                                  & ESS9                                   & CFPS                                   & WVS                                    & \multicolumn{1}{c|}{CGSS}                                   & Avg.                                    & ESS11                                  & ESS9                                   & CFPS                                   & WVS                                    & \multicolumn{1}{c|}{CGSS}                                   & Avg.                                    \\ \midrule
-                           & TS                          & 0.130                                  & 0.137                                  & 0.141                                  & 0.144                                  & \multicolumn{1}{c|}{0.123}                                  & 0.135                                  & 0.012                                  & 0.011                                  & 0.008                                  & 0.015                                  & \multicolumn{1}{c|}{0.014}                                  & 0.012                                  \\ \midrule
                            & Direct                      & 0.799                                  & 0.773                                  & 0.899                                  & 0.844                                  & \multicolumn{1}{c|}{0.723}                                  & 0.808                                  & 0.141                                  & 0.129                                  & 0.328                                  & 0.203                                  & \multicolumn{1}{c|}{0.258}                                  & 0.212                                  \\
                            & PE                          & 0.715                                  & 0.803                                  & 0.872                                  & 0.801                                  & \multicolumn{1}{c|}{0.738}                                  & 0.786                                  & 0.135                                  & 0.124                                  & 0.314                                  & 0.214                                  & \multicolumn{1}{c|}{0.246}                                  & 0.207                                  \\
                            & AAE                         & 0.125                                  & 0.132                                  & 0.127                                  & 0.134                                  & \multicolumn{1}{c|}{0.117}                                  & 0.127                                  & 0.011                                  & 0.012                                  & 0.014                                  & 0.015                                  & \multicolumn{1}{c|}{\textbf{0.008}}                         & 0.012                                  \\
                            & TKFT                        & 0.119                                  & 0.128                                  & 0.118                                  & 0.127                                  & \multicolumn{1}{c|}{0.115}                                  & 0.121                                  & 0.012                                  & 0.011                                  & 0.014                                  & 0.015                                  & \multicolumn{1}{c|}{\textbf{0.008}}                         & 0.012                                  \\
\multirow{-5}{*}{Qwen3-4B}  & \cellcolor[HTML]{EFEFEF}DSA & \cellcolor[HTML]{EFEFEF}\textbf{0.108} & \cellcolor[HTML]{EFEFEF}\textbf{0.111} & \cellcolor[HTML]{EFEFEF}\textbf{0.095} & \cellcolor[HTML]{EFEFEF}\textbf{0.109} & \multicolumn{1}{c|}{\cellcolor[HTML]{EFEFEF}\textbf{0.106}} & \cellcolor[HTML]{EFEFEF}\textbf{0.106} & \cellcolor[HTML]{EFEFEF}\textbf{0.008} & \cellcolor[HTML]{EFEFEF}\textbf{0.007} & \cellcolor[HTML]{EFEFEF}\textbf{0.008} & \cellcolor[HTML]{EFEFEF}\textbf{0.012} & \multicolumn{1}{c|}{\cellcolor[HTML]{EFEFEF}\textbf{0.008}} & \cellcolor[HTML]{EFEFEF}\textbf{0.009} \\ \midrule
                            & Direct                      & 1.413                                  & 0.929                                  & 0.778                                  & 0.825                                  & \multicolumn{1}{c|}{0.766}                                  & 0.942                                  & 0.248                                  & 0.161                                  & 0.286                                  & 0.214                                  & \multicolumn{1}{c|}{0.283}                                  & 0.238                                  \\
                            & PE                          & 1.252                                  & 1.078                                  & 0.851                                  & 0.766                                  & \multicolumn{1}{c|}{0.841}                                  & 0.958                                  & 0.237                                  & 0.171                                  & 0.297                                  & 0.195                                  & \multicolumn{1}{c|}{0.270}                                  & 0.234                                  \\
                            & AAE                         & 0.126                                  & 0.139                                  & 0.120                                  & 0.130                                  & \multicolumn{1}{c|}{0.129}                                  & 0.129                                  & 0.011                                  & 0.011                                  & 0.013                                  & 0.014                                  & \multicolumn{1}{c|}{0.009}                                  & 0.012                                  \\
                            & TKFT                        & 0.113                                  & 0.128                                  & 0.117                                  & 0.122                                  & \multicolumn{1}{c|}{0.113}                                  & 0.119                                  & 0.012                                  & 0.011                                  & 0.014                                  & 0.014                                  & \multicolumn{1}{c|}{\textbf{0.008}}                         & 0.012                                  \\
\multirow{-5}{*}{Qwen3-32B} & \cellcolor[HTML]{EFEFEF}DSA & \cellcolor[HTML]{EFEFEF}\textbf{0.105} & \cellcolor[HTML]{EFEFEF}\textbf{0.111} & \cellcolor[HTML]{EFEFEF}\textbf{0.097} & \cellcolor[HTML]{EFEFEF}\textbf{0.102} & \multicolumn{1}{c|}{\cellcolor[HTML]{EFEFEF}\textbf{0.102}} & \cellcolor[HTML]{EFEFEF}\textbf{0.103} & \cellcolor[HTML]{EFEFEF}\textbf{0.008} & \cellcolor[HTML]{EFEFEF}\textbf{0.007} & \cellcolor[HTML]{EFEFEF}\textbf{0.008} & \cellcolor[HTML]{EFEFEF}\textbf{0.011} & \multicolumn{1}{c|}{\cellcolor[HTML]{EFEFEF}\textbf{0.008}} & \cellcolor[HTML]{EFEFEF}\textbf{0.008} \\ \midrule
                            & Direct                      & 1.041                                  & 1.101                                  & 0.998                                  & 0.772                                  & \multicolumn{1}{c|}{0.843}                                  & 0.951                                  & 0.177                                  & 0.184                                  & 0.350                                  & 0.191                                  & \multicolumn{1}{c|}{0.296}                                  & 0.240                                  \\
                            & PE                          & 0.900                                  & 1.179                                  & 0.886                                  & 0.670                                  & \multicolumn{1}{c|}{0.760}                                  & 0.879                                  & 0.173                                  & 0.178                                  & 0.334                                  & 0.177                                  & \multicolumn{1}{c|}{0.257}                                  & 0.224                                  \\
                            & AAE                         & 0.133                                  & 0.191                                  & 0.207                                  & 0.154                                  & \multicolumn{1}{c|}{0.135}                                  & 0.164                                  & 0.012                                  & 0.017                                  & 0.023                                  & 0.017                                  & \multicolumn{1}{c|}{0.009}                                  & 0.016                                  \\
                            & TKFT                        & 0.126                                  & 0.167                                  & 0.154                                  & 0.126                                  & \multicolumn{1}{c|}{0.124}                                  & 0.139                                  & 0.011                                  & 0.014                                  & 0.019                                  & 0.015                                  & \multicolumn{1}{c|}{\textbf{0.008}}                         & 0.013                                  \\
\multirow{-5}{*}{Llama3-8B} & \cellcolor[HTML]{EFEFEF}DSA & \cellcolor[HTML]{EFEFEF}\textbf{0.104} & \cellcolor[HTML]{EFEFEF}\textbf{0.139} & \cellcolor[HTML]{EFEFEF}\textbf{0.133} & \cellcolor[HTML]{EFEFEF}\textbf{0.086} & \multicolumn{1}{c|}{\cellcolor[HTML]{EFEFEF}\textbf{0.101}} & \cellcolor[HTML]{EFEFEF}\textbf{0.113} & \cellcolor[HTML]{EFEFEF}\textbf{0.008} & \cellcolor[HTML]{EFEFEF}\textbf{0.009} & \cellcolor[HTML]{EFEFEF}\textbf{0.011} & \cellcolor[HTML]{EFEFEF}\textbf{0.010} & \multicolumn{1}{c|}{\cellcolor[HTML]{EFEFEF}\textbf{0.008}} & \cellcolor[HTML]{EFEFEF}\textbf{0.009} \\ \bottomrule
\end{tabular}
}
\caption{Comparison of different methods on the enhancement setting.}
\label{tab:simulation}
\end{table*}

\section{Experiment}

\subsection{Experimental Setup}

\noindent\textbf{Datasets:}
Following the dataset construction workflow of OpinionQA \cite{opinionqa}, we build our benchmarks by extracting core questions and associated background questions from five representative survey datasets spanning diverse regions, languages, and domains: ESS11, ESS9 (European Social Survey) \cite{ess}, WVS (World Values Survey) \cite{wvs_dataset}, CGSS (Chinese General Social Survey) \cite{cgss_survey} and CFPS (China Family Panel Studies) \cite{CFPS}.

\paragraph{Enhancement.}
For the task of enhancement, we select 20 core questions from each dataset to be simulated. For each core question, we include 5--6 background questions that reflect respondents' characteristics (e.g., gender, age, political orientation). To simulate a realistic low-budget survey setting, we randomly sample data from each dataset as the training set, and regard the overall distribution of the full dataset as the ground-truth distribution to be simulated. All reported results are averaged over five independent random training-set samples.

\paragraph{Generation.}
To enable direct comparison with prior work, we adopt the dataset from \cite{TKFT2}. Specifically, a subset of WVS was used as the training set, and evaluate on three held-out test sets: (i) \textbf{WVS-NewQ} (C1-Q3 in original paper), which contains unseen questions; (ii) \textbf{WVS-NewP} (C2-Q1), which contains unseen populations; and (iii) \textbf{WVS-Both} (C2-Q3), where both the questions and the populations differ from training.

\noindent\textbf{Prompts:} We use the original survey questions and responses as prompts. In addition, we append an instruction at the end of each prompt, asking the LLM to directly output the tokens corresponding to the predefined options. A detailed example of our constructed prompt is provided in \Cref{sec:appendixprompt}.

\noindent\textbf{LLMs:} We employ two variants of Qwen3 \cite{qwen3}: Qwen3-4B and Qwen3-32B---and additionally include Llama3-8B \cite{llama3} in our experiments. They are widely used as strong baselines in recent research.

\noindent\textbf{Implementation details.} In Stage 2, we enumerate background pairs with Hamming distance 1 and keep only pairs where both subgroups contain at least 10 respondents. Mini-batches are sampled from this filtered pair set, and gradients are backpropagated only through the smaller subgroup prediction. All Enhancement results are averaged over five random calibration samples; all methods are trained for 1 epoch in Enhancement.

\noindent\textbf{Metrics:} Since one-hot accuracy is unsuitable for evaluating simulated distributions, we report Jensen--Shannon divergence (JSD) and Wasserstein-1 distance (WD). Let $p$ and $q$ denote the predicted and reference distributions, respectively. The JSD is defined as
\begin{equation}
\mathrm{JSD}(p,q)
=
\frac{1}{2}\mathrm{KL}(p\|m)
+
\frac{1}{2}\mathrm{KL}(q\|m),
\qquad
m=\frac{1}{2}(p+q),
\end{equation}
and the WD is defined as
\begin{equation}
W_1(p,q)
=
\min_{\gamma \in \Pi(p,q)}
\sum_{i,j}\gamma_{ij} d(i,j),
\end{equation}
where $\Pi(p,q)$ is the set of couplings between $p$ and $q$, and $d(i,j)$ is the ground distance between options $i$ and $j$. Lower values indicate better agreement.

\subsection{Baselines}

\noindent\textbf{Training Set as Result (TS):} One of the key baselines is directly using the distribution of training set as the predicted result. This baseline is often ignored in previous research, but it provides a simple yet effective benchmark to assess the performance of fine-tuning methods.

\noindent\textbf{Zero-shot Methods:} 

\noindent\textbf{Direct:} Directly prompting the LLM to generate the results without any fine-tuning or prompt engineering \cite{opinionqa}.

\noindent\textbf{Prompt Engineering (PE):} Using a designed prompt to better simulate the respondents' personality and preferences \cite{PE}.

\noindent\textbf{Fine-tuning Methods:}  

\noindent\textbf{AAE:} AAE fine-tunes a lightweight conditional probability model that maps the LLM’s output distribution to the distribution of the training set \cite{AAE}.

\noindent\textbf{Token-Level Fine-tuning (TKFT):} TKFT directly fine-tunes the token-level output probabilities of the LLM to align with the training set \cite{TKFT1}.

\subsection{Enhancement Results}

\begin{table}[]
\centering
\resizebox{.99\linewidth}{!}{
\begin{tabular}{@{}l|l|cccc|cccc@{}}
\toprule
                            &                             & \multicolumn{4}{c|}{WD$\downarrow$}                                                                                                                                                                & \multicolumn{4}{c}{JSD$\downarrow$}                                                                                                                                                                \\ \cmidrule(l){3-10} 
\multirow{-2}{*}{LLM}       & \multirow{-2}{*}{Method}    & NewQ                                   & NewP                                   & \multicolumn{1}{c|}{Both}                                   & Avg.                                   & NewQ                                   & NewP                                   & \multicolumn{1}{c|}{Both}                                   & Avg.                                   \\ \midrule
                            & Direct                      & 0.084                                  & 0.095                                  & \multicolumn{1}{c|}{0.138}                                  & 0.106                                  & 0.417                                  & 0.355                                  & \multicolumn{1}{c|}{0.361}                                  & 0.378                                  \\
                            & PE                          & 0.086                                  & 0.091                                  & \multicolumn{1}{c|}{0.137}                                  & 0.105                                  & 0.421                                  & 0.354                                  & \multicolumn{1}{c|}{0.351}                                  & 0.376                                  \\
                            & TKFT                        & 0.073                                  & 0.076                                  & \multicolumn{1}{c|}{0.077}                                  & 0.075                                  & 0.244                                  & 0.219                                  & \multicolumn{1}{c|}{0.197}                                  & 0.220                                  \\
\multirow{-4}{*}{Qwen3-4B}  & \cellcolor[HTML]{EFEFEF}DSA & \cellcolor[HTML]{EFEFEF}\textbf{0.058} & \cellcolor[HTML]{EFEFEF}\textbf{0.067} & \multicolumn{1}{c|}{\cellcolor[HTML]{EFEFEF}\textbf{0.071}} & \cellcolor[HTML]{EFEFEF}\textbf{0.065} & \cellcolor[HTML]{EFEFEF}\textbf{0.226} & \cellcolor[HTML]{EFEFEF}\textbf{0.208} & \multicolumn{1}{c|}{\cellcolor[HTML]{EFEFEF}\textbf{0.191}} & \cellcolor[HTML]{EFEFEF}\textbf{0.208} \\ \midrule
                            & Direct                      & 0.088                                  & 0.105                                  & \multicolumn{1}{c|}{0.101}                                  & 0.098                                  & 0.342                                  & 0.317                                  & \multicolumn{1}{c|}{0.281}                                  & 0.313                                  \\
                            & PE                          & 0.091                                  & 0.103                                  & \multicolumn{1}{c|}{0.104}                                  & 0.099                                  & 0.368                                  & 0.307                                  & \multicolumn{1}{c|}{0.292}                                  & 0.322                                  \\
                            & TKFT                        & 0.062                                  & 0.067                                  & \multicolumn{1}{c|}{0.073}                                  & 0.067                                  & 0.200                                  & 0.179                                  & \multicolumn{1}{c|}{0.185}                                  & 0.188                                  \\
\multirow{-4}{*}{Qwen3-32B} & \cellcolor[HTML]{EFEFEF}DSA & \cellcolor[HTML]{EFEFEF}\textbf{0.051} & \cellcolor[HTML]{EFEFEF}\textbf{0.062} & \multicolumn{1}{c|}{\cellcolor[HTML]{EFEFEF}\textbf{0.069}} & \cellcolor[HTML]{EFEFEF}\textbf{0.061} & \cellcolor[HTML]{EFEFEF}\textbf{0.183} & \cellcolor[HTML]{EFEFEF}\textbf{0.169} & \multicolumn{1}{c|}{\cellcolor[HTML]{EFEFEF}\textbf{0.18}}  & \cellcolor[HTML]{EFEFEF}\textbf{0.177} \\ \midrule
                            & Direct                      & 0.097                                  & 0.116                                  & \multicolumn{1}{c|}{0.097}                                  & 0.103                                  & 0.251                                  & 0.232                                  & \multicolumn{1}{c|}{0.241}                                  & 0.241                                  \\
                            & PE                          & 0.100                                  & 0.114                                  & \multicolumn{1}{c|}{0.094}                                  & 0.103                                  & 0.267                                  & 0.212                                  & \multicolumn{1}{c|}{0.232}                                  & 0.237                                  \\
                            & TKFT                        & 0.081                                  & 0.073                                  & \multicolumn{1}{c|}{0.087}                                  & 0.080                                  & 0.230                                  & 0.142                                  & \multicolumn{1}{c|}{0.227}                                  & 0.200                                  \\
\multirow{-4}{*}{Llama3-8B} & \cellcolor[HTML]{EFEFEF}DSA & \cellcolor[HTML]{EFEFEF}\textbf{0.064} & \cellcolor[HTML]{EFEFEF}\textbf{0.064} & \multicolumn{1}{c|}{\cellcolor[HTML]{EFEFEF}\textbf{0.079}} & \cellcolor[HTML]{EFEFEF}\textbf{0.069} & \cellcolor[HTML]{EFEFEF}\textbf{0.207} & \cellcolor[HTML]{EFEFEF}\textbf{0.133} & \multicolumn{1}{c|}{\cellcolor[HTML]{EFEFEF}\textbf{0.217}} & \cellcolor[HTML]{EFEFEF}\textbf{0.186} \\ \bottomrule
\end{tabular}
}
\caption{Comparison of different methods on the generation setting. \emph{NewQ} tests on new questions, \emph{NewP} tests on new populations, and \emph{Both} changes both.}
\label{tab:generation}
\end{table}

\Cref{tab:simulation} reports the performance of different methods under the Enhancement setting, where we fine-tune on a small training set for each target question and evaluate on held-out responses from the same question. DSA consistently achieves the best performance across all datasets and backbone LLMs, clearly outperforming both zero-shot and fine-tuning baselines. Zero-shot approaches exhibit strong distributional bias, indicating that prompting an LLM without adaptation is insufficient for accurate distribution estimation. Although fine-tuning baselines improve over zero-shot, their estimates typically remain close to the empirical training-set baseline and provide limited gains beyond it. In contrast, DSA’s distribution shift alignment leverages background-driven shift supervision during calibration, producing distributions closer to the held-out ground-truth distribution. In addition, DSA is the only method that significantly outperforms the TS baseline across all datasets, indicating that it is not merely fitting the training data but instead leveraging the shift to generalize. An example of the output distribution is provided in \Cref{sec:distributioncase}.

While the larger model typically achieves lower divergence after calibration, this gain is not observed in zero-shot. This indicates that scaling helps mainly when paired with adaptation, and that calibration remains crucial for aligning LLM outputs with human-like choices.

\subsection{Generation Results}

\Cref{tab:generation} summarizes results in the Generation setting, where models are trained once and directly evaluated on unseen test splits without any further adaptation. DSA achieves the best overall performance across all backbones and all splits. This indicates that capturing preference shifts across backgrounds can generalize to unseen questions and even new populations, improving distribution simulation beyond directly fitting absolute distributions.

Among the three splits, DSA brings the largest gains on new questions with the same population, while the improvement is more limited on new population. This can be explained by that the shift is learned from the training population. If the population changes, the learned shifts may no longer match, leading to smaller gains.

\begin{figure}[t]
\centering
  \includegraphics[width=0.49\linewidth]{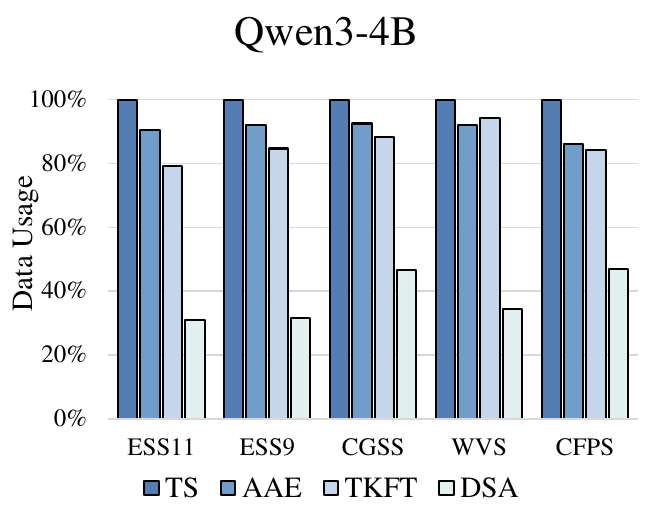}
  \includegraphics[width=0.49\linewidth]{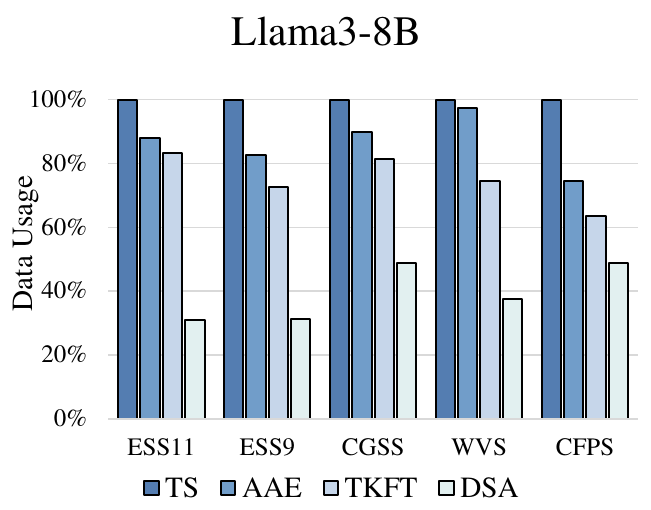}
  \caption {Comparison of required real survey data to reach similar accuracy across different methods, reflecting the relative data efficiency of different approaches.}
    \label{fig:data}
\end{figure}

\subsection{Data Efficiency and Cost Reduction}

An important indicator of whether a method can effectively enhance real survey result is how much survey data it can save, thereby reducing the cost of conducting surveys. This comparison is often ignored in previous studies. To evaluate this, we compare the data efficiency of different methods by estimating how much training data can be saved while maintaining a similar accuracy. Specifically, we measure how many training samples different fine-tuning methods would need to reach similar performance, as shown in \Cref{fig:data}. For completeness, the results with Qwen3-32B are reported in the appendix.

While AAE and TKFT achieve certain improvements, their data efficiency gains are limited. In contrast, DSA consistently reaches comparable accuracy while saving at least 53.48\% of the training data, even with an LLM with only 4B parameters. This advantage stems from the second-stage alignment of distribution shifts across backgrounds, which allows the model to generalize learned relations beyond observed samples and thus achieve more accurate distribution than the training set.

\begin{figure*}[t]
\centering
\begin{subfigure}{0.24\linewidth}
    \includegraphics[width=\linewidth]{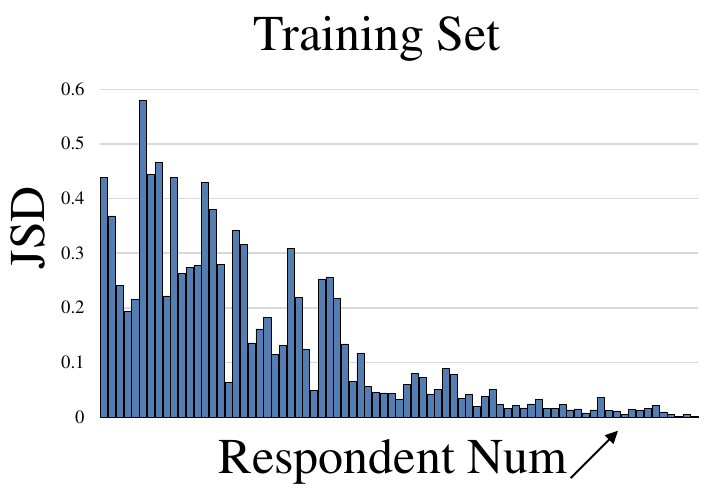}
\end{subfigure}
\begin{subfigure}{0.24\linewidth}
    \includegraphics[width=\linewidth]{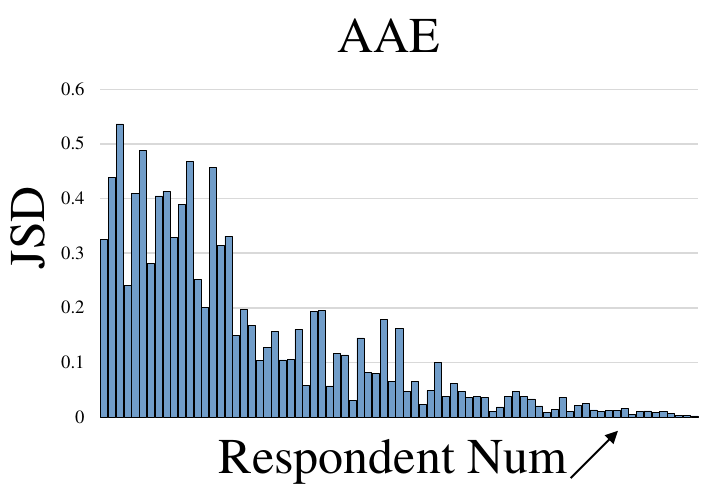}
\end{subfigure}
\begin{subfigure}{0.24\linewidth}
    \includegraphics[width=\linewidth]{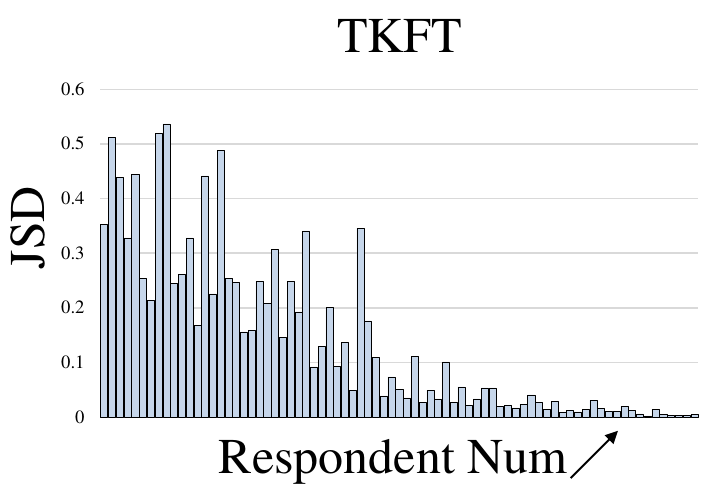}
\end{subfigure}
\begin{subfigure}{0.24\linewidth}
    \includegraphics[width=\linewidth]{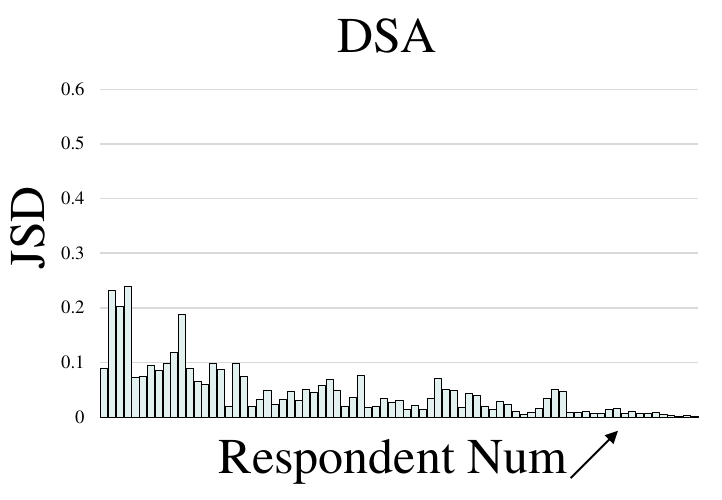}
\end{subfigure}

\vspace{2mm}

\begin{subfigure}{0.24\linewidth}
    \includegraphics[width=\linewidth]{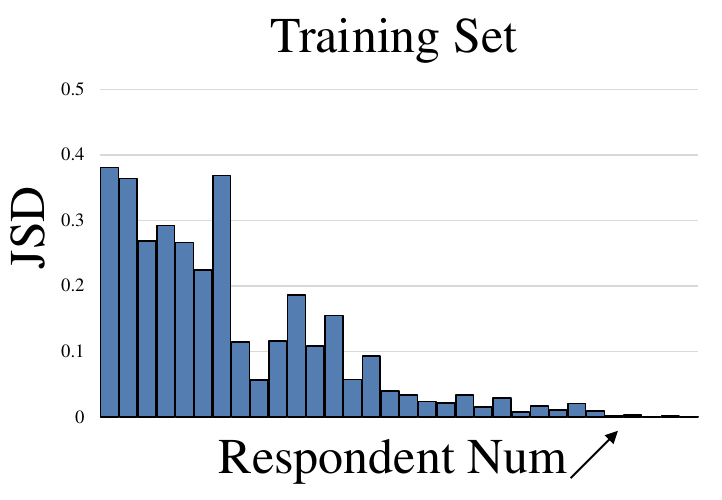}
\end{subfigure}
\begin{subfigure}{0.24\linewidth}
    \includegraphics[width=\linewidth]{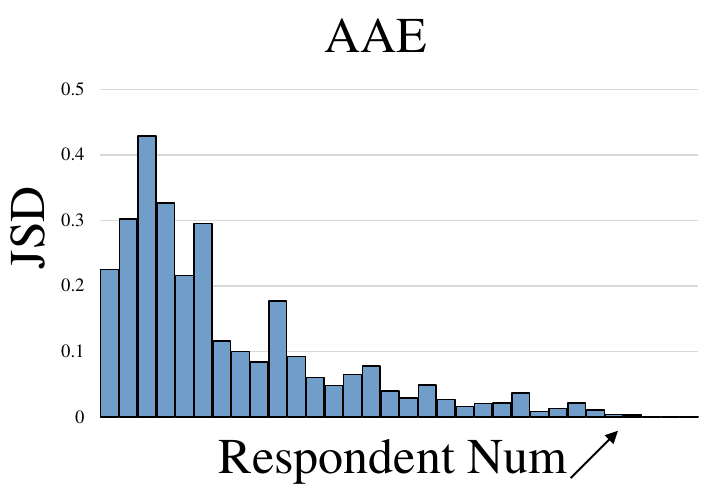}
\end{subfigure}
\begin{subfigure}{0.24\linewidth}
    \includegraphics[width=\linewidth]{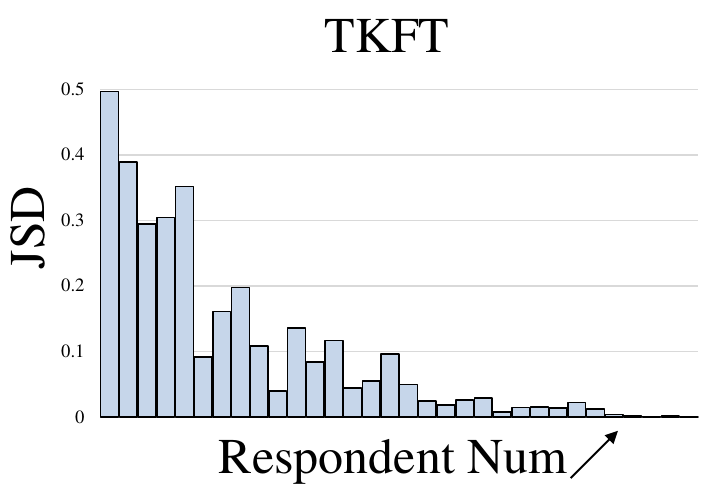}
\end{subfigure}
\begin{subfigure}{0.24\linewidth}
    \includegraphics[width=\linewidth]{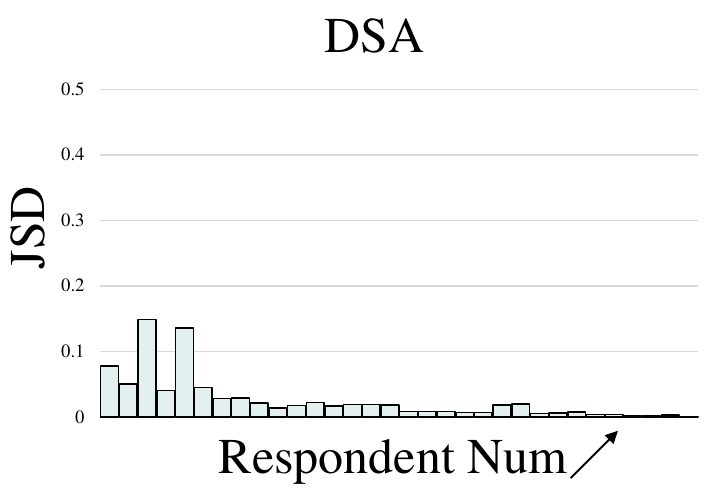}
\end{subfigure}

\caption{JSD of different methods across backgrounds in ESS11. Backgrounds are sorted from left to right by the number of respondents, with rarer backgrounds appearing on the left.}
\label{fig:cases}
\end{figure*}

\begin{figure}[t]
\centering
  \includegraphics[width=0.99\linewidth]{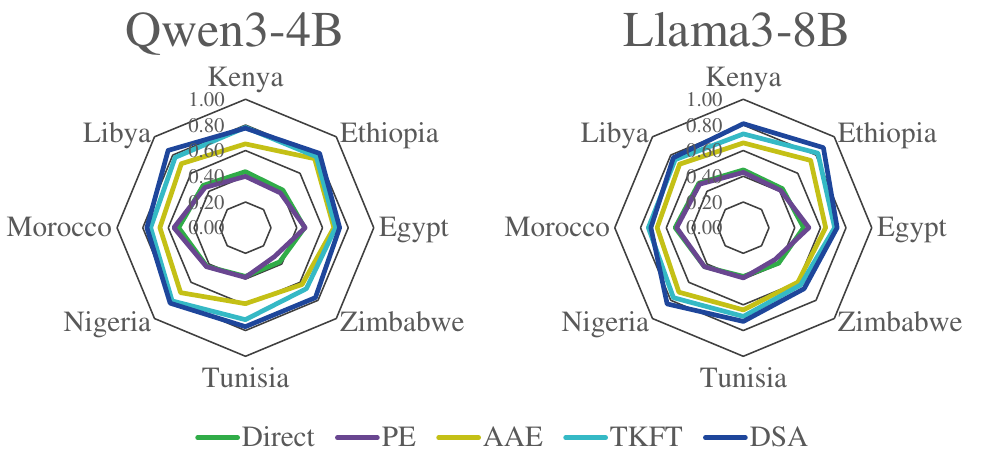}
  \caption {Accuracy of different methods on population from unseen countries.}
    \label{fig:country}
\end{figure}

\subsection{Generalization across Different Backgrounds}

In real-world surveys, some background groups are hard to sample, leading to missing or even unseen subgroups in the training data. We evaluate generalization to such unseen backgrounds on two ESS11 cases by sorting backgrounds by respondent count and reporting per-background divergence (JSD). As shown in \Cref{fig:cases}, DSA substantially reduces JSD for rare groups and remains accurate even when a background is entirely absent from the training set. 

We further confirm this advantage in the Generation setting using the same unseen-background evaluation as \cite{TKFT2}. Notably, even without target-domain adaptation, DSA achieves higher accuracy on unseen backgrounds across different countries.

\begin{table}[t]
\centering
\resizebox{\linewidth}{!}{
\begin{tabular}{lccccc}
\hline
Method & ESS11            & ESS9             & CGSS             & WVS              & CFPS             \\ \hline
Direct & \textcolor{white}{0}7.36\%            & \textcolor{white}{0}6.91\%            & \textcolor{white}{0}9.58\%            & 11.70\%           & \textcolor{white}{0}6.21\%            \\
PE     & \textcolor{white}{0}9.22\%            & \textcolor{white}{0}8.15\%            & 13.40\%           & 12.31\%           & 10.57\%           \\
AAE    & 59.26\%          & 52.40\%          & 49.44\%          & 60.94\%          & 48.98\%          \\
TKFT   & 54.94\%          & 48.40\%          & 54.44\%          & 57.81\%          & 63.27\%          \\
DSA    & \textbf{98.77\%} & \textbf{96.40\%} & \textbf{96.11\%} & \textbf{97.27\%} & \textbf{93.88\%} \\ \hline
\end{tabular}
}
\caption{Proportion of backgrounds where each method produces a distribution closer to the true one than the training set.}
\label{tab:robustquestion}
\end{table}

\subsection{Robustness across Questions}

When simulating survey data, robustness across different questions is crucial, as the true underlying distribution in reality is unobservable. Assuming we have a small amount of real survey data and use a specific method to simulate additional data, we cannot directly verify whether this method actually brings the simulated distribution closer to the true underlying distribution. We calculate the proportion of background groups for which a method produces a distribution closer to the true one than the training set itself. This metric reflects how reliably each method improves upon the observed data.

As shown in \Cref{tab:robustquestion}, the proposed DSA achieves improvements on more than 90\% of backgrounds across all datasets, indicating that it carries a very low risk of performance degradation when applied to real survey data. In contrast, the other two fine-tuning methods, AAE and TKFT, achieve moderate improvements, with the best reaching 63.27\%.

\begin{figure*}[t]
\centering
  \includegraphics[width=0.9\linewidth]{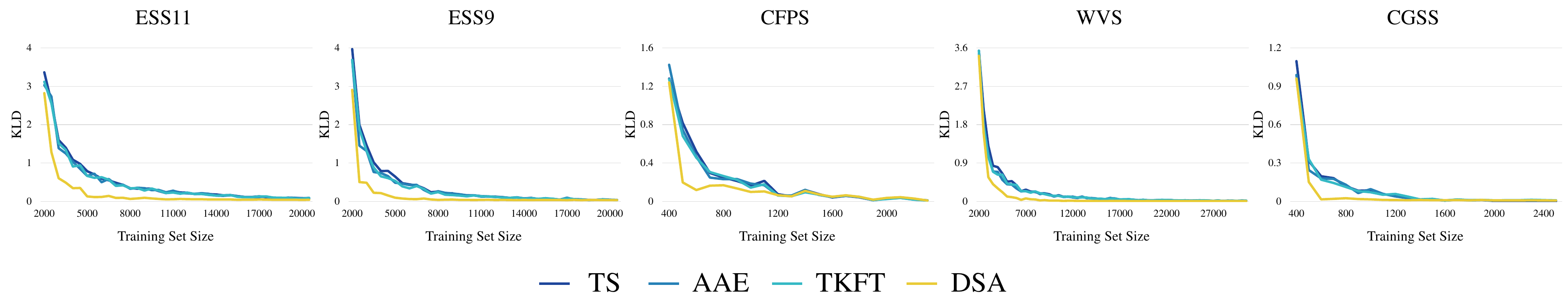}
  \caption {KL divergence of different fine-tuning methods across different training set sizes.}
    \label{fig:trainset}
\end{figure*}

\subsection{Robustness across Training Set Size}

In real-world applications, surveys vary greatly in scale, and the amount of available training data can differ significantly. Therefore, it is also important whether a fine-tuning method can consistently improve performance across different training set sizes. In this experiment, we evaluate various methods by gradually increasing the amount of real survey data used for fine-tuning and measuring their performance under each setting. As shown in \Cref{fig:trainset}, DSA provides most stable and consistent improvements across most of the training set sizes. However, when the training set becomes sufficiently large, the improvements brought by different methods become marginal, suggesting that the primary advantage of LLM-based survey simulation lies in medium-scale surveys, where real survey data are limited but not extremely scarce.

\begin{table}[t]
\centering
\resizebox{\linewidth}{!}{
\begin{tabular}{lccccc}
\hline
Method & ESS11         & ESS9          & CGSS          & WVS           & CFPS          \\ \hline
Direct & 1.57          & 1.07          & 0.85          & 1.82          & 1.01          \\
PE     & 1.13          & 1.52          & 1.30           & 2.45          & 1.14          \\
AAE    & 0.17          & 0.22          & 0.13          & \textbf{0.06}          & 0.17          \\
TKFT   & 0.09          & 0.16          & \textbf{0.04}          & 0.11          & 0.07          \\
DSA    & \textbf{0.06} & \textbf{0.11} & 0.08 & 0.07 & \textbf{0.05} \\ \hline
\end{tabular}
}
\caption{Average pairwise $\text{JSD} \times 100$ between outputs generated from semantically equivalent but differently phrased prompts. Lower values indicate stronger consistency across prompts.}
\label{tab:robustprompt}
\end{table}

\subsection{Consistency across Prompts}

A robust simulation method should be insensitive to prompt, as changes in linguistic expression should not alter the underlying prediction. To evaluate this property, we generate four additional sets of semantically equivalent but syntactically varied prompts using GPT-5 for each dataset, and test how consistent each method’s predicted distributions remain across these prompts. \Cref{tab:robustprompt} reports the average pairwise JSD among outputs from different prompts.

As shown in the results, DSA exhibits strong consistency across most questions, indicating its robustness to prompt variation. Zero-shot methods, in contrast, are highly sensitive to prompt wording, with substantial shifts in predicted distributions when the phrasing changes. Fine-tuning methods demonstrate greater stability, as training aligns the model’s internal representations with the survey context.

\begin{table}[t]
\centering
\resizebox{\linewidth}{!}{
\begin{tabular}{@{}ccccc@{}}
\toprule
\multirow{2}{*}{LLM} & \multicolumn{2}{c}{Fine-tuning} & \multicolumn{2}{c}{Task} \\
                     & Phase        & Method           & Enhancement & Generation \\ \midrule
4B                   & 1            & Full             & 0.118       & 0.073      \\
4B                   & 1+2          & Full             & 0.107       & 0.055      \\
4B                   & 1+2          & Lora             & 0.111       & 0.064      \\
4B                   & 1+2          & LastLayer        & 0.108       & 0.058      \\
32B                  & 1+2          & LastLayer        & 0.105       & 0.051      \\ \bottomrule
\end{tabular}
}
\caption{WD of different fine-tuning phases, strategies, and model sizes on both tasks.}
\label{tab:ablation}
\end{table}

\subsection{Ablation Study}

To further examine the contribution of each component and compare different fine-tuning strategies, we evaluate the WD of different fine-tuning phases, model sizes and fine-tuning strategies on both tasks, as summarized in Table~\ref{tab:ablation}. The results show that the second-phase of distribution shift alignment reduces the error, confirming the effectiveness of the proposed shift alignment loss. Comparing different fine-tuning strategies, updating only the last transformer layer achieves performance comparable to full-model fine-tuning. Furthermore, model size matters more in the generation task, which relies on generalization ability of the LLM, while enhancement benefits less from scaling given its question-specific adaptation.

\section{Related Work}

\subsection{LLM-Based Simulation}

Recent studies have explored whether LLMs can serve as proxies for human respondents in social or market surveys. A series of works investigate zero-shot simulation, where LLMs generate answers to survey questions directly, without fine-tuning on human data \cite{pmlr-v202-aher23a, zeroshot1, zeroshot2}. For instance, \cite{zeroshot} shows that LLMs conditioned on demographic backgrounds can generate responses that closely match human subpopulation averages. In addition, prompt engineering has been widely adopted to improve zero-shot performance. By carefully crafting prompts with specific instructions or background context, researchers have shown that LLMs can better align with human responses across diverse survey questions \cite{PE, PE1}. However, several studies \cite{senitivity, senitivity1} find that these zero-shot methods are highly sensitive to prompt design and randomness. Additionally, such methods often underestimate the variance of human opinions \cite{nature}.

Recent research proposed to fine-tune LLMs to explicitly align their outputs with survey data distributions. For example, the AAE \cite{AAE} fine-tunes an additional probabilistic conditional model to map the distribution of LLM outputs to that of the training set. Other studies \cite{TKFT1, TKFT2} directly fine-tune LLMs on human survey data, aligning their token-level probabilities with the training set's distribution.

\subsection{Distributional Preference Modeling and Alignment}

An emerging research direction explores aligning large language models with population-level human preferences rather than individual judgments. Recent works on pluralistic or distributional alignment \cite{alignment1,alignment2,alignment3,alignment4} aim to make LLMs sensitive to diverse value systems and group-level differences.

Our work differs from previous approaches in that we focus on modeling distributional shifts rather than directly fitting observed survey distributions. Our method leverages the LLM’s ability to capture relative differences in preferences across backgrounds.

\section{Conclusion}

We proposed Distribution Shift Alignment (DSA), a fine-tuning framework that leverages LLMs’ ability to capture preference shifts across backgrounds. By aligning both distributions and their shifts, DSA produces estimates closer to the true distribution than the training data. Experiments on five survey datasets show that DSA consistently outperforms existing methods while reducing real data requirements by up to 69\%.

\clearpage

\section*{Limitations}

Although DSA demonstrates strong performance in simulating survey distributions, it still has several limitations. Our goal is not to recover human preferences in an absolute sense, but to estimate what a larger survey conducted under the same questionnaire design would have produced from a limited sample.

\noindent\textbf{Dependence on independence assumptions.} 
Our theoretical analysis relies on the assumption that background variables are independent. This assumption is introduced to provide theoretical intuition for why shift-based estimation can reduce reliance on noisy subgroup empirical distributions, rather than as an algorithmic condition required by DSA. While this simplifies the estimation of distribution shifts, real-world survey variables are often correlated. Violations of this assumption may reduce the accuracy of the shift-based alignment.

Empirically, however, DSA remains effective even when background variables are correlated, which may be because LLM attention implicitly captures such dependencies, and/or because shift alignment introduces a regularizing effect that improves robustness. These possibilities suggest promising directions for future work, including explicit joint modeling of correlated backgrounds and a deeper study of why DSA is resilient to dependence.

\noindent\textbf{Limited evaluation scope.} 
Our experiments focus on datasets with discrete, well-structured survey questions. The method’s effectiveness in open-ended, multi-choice, or free-text survey settings remains untested. Extending DSA to more complex or less-structured survey formats would further validate its general applicability.

A potential risk of our method is that the simulated distributions may inadvertently amplify existing biases in the training data or in the language model itself, leading to distorted representations of certain demographic groups or opinions.

In the experiment, we use LLMs to assist in filtering and selecting survey data, enabling more efficient and targeted dataset construction for simulation and analysis.

\section*{Acknowledgments}
We thank the anonymous reviewers for their careful reading and constructive feedback, which helped us improve the clarity, motivation, and presentation of this paper. Mengfei Li was supported by the National Natural Science Foundation of China (grants No.724B2010; No.72293560/72293565). Ji Huang and Shuai Shao were supported by the National Natural Science Foundation of China (No. 62572452).

\bibliography{custom}

\clearpage

\appendix

\begin{figure}[t]
\centering
  \includegraphics[width=0.95\linewidth]{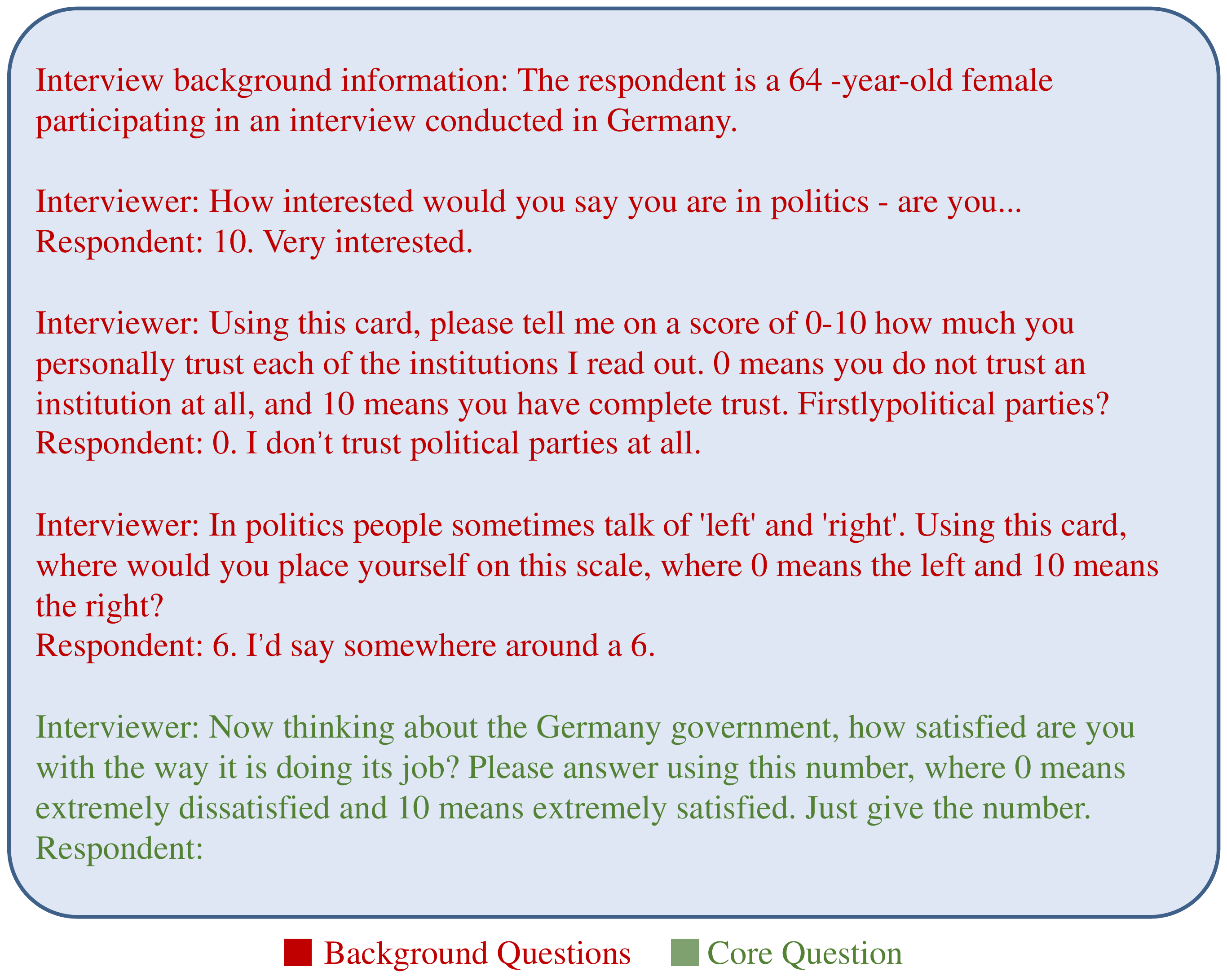}
  \caption {An example of constructed sample from ESS.}
    \label{fig:prompt}
\end{figure}

\begin{figure*}[t]
\centering
  \includegraphics[width=0.95\linewidth]{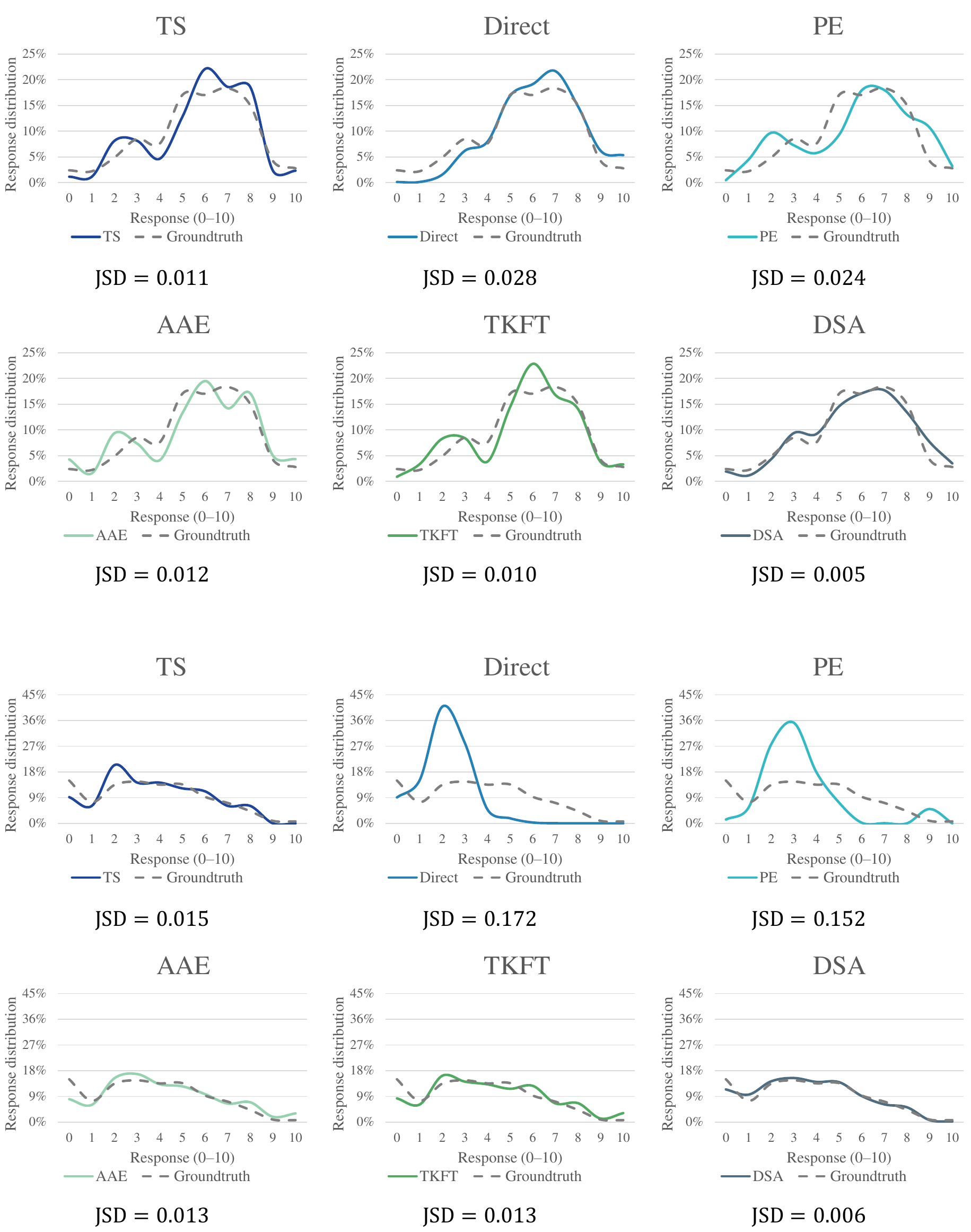}
  \caption {Examples of output distribution from ESS.}
    \label{fig:distributioncase}
\end{figure*}

\section{Distribution Case}\label{sec:distributioncase}

\Cref{fig:distributioncase} provides two representative ESS11 examples illustrating how different methods predict the full response distribution. 
In each panel, the dashed gray curve is the empirical ground-truth distribution computed from the full dataset, while the colored curve is the distribution predicted by a given method using only the small calibration (training) set. We report the Jensen--Shannon divergence (JSD; lower is better) under each panel.

Both examples in \Cref{fig:distributioncase} correspond to small subgroups with limited numbers of respondents, where the sampled training set can deviate noticeably from the full-dataset (ground-truth) distribution. 
While baseline fine-tuning strategies can partially leverage the LLM’s generalization to correct this sampling-induced shift, their predicted distributions still exhibit visible bias relative to the ground truth. 
In contrast, DSA more consistently realigns the predicted distribution toward the full-dataset target, yielding the smallest divergences in both cases (JSD $=0.005$ and $0.006$), with particularly clear improvements when the training set is strongly shifted.

\section{Prompt}
\label{sec:appendixprompt}

\Cref{fig:prompt} illustrates a representative prompt for simulating survey responses in ESS11. 
Items highlighted in \textcolor{red}{red} are \emph{background questions} used to condition the simulation. 
In this example, they include demographics (a 64-year-old female interviewed in Germany) and political predispositions such as interest in politics (10/10), trust in political parties (0/10), and left--right self-placement (6/10). 
The \textcolor{green}{green} item is the \emph{core question} to be simulated—satisfaction with the German government on a 0--10 scale—which is naturally related to these background attributes as they provide relevant context about political engagement, ideology, and institutional trust. 
We define a \emph{subgroup} by an identical tuple of background answers (i.e., the red items); respondents sharing the same tuple are grouped together and are assigned the same filled-in prompt. 
This prompt uses the original questionnaire wording and responses from ESS11. In our setup, a \emph{subgroup} is defined by a unique combination of background-question answers (i.e., respondents sharing the same values on the red-highlighted items). 
For each subgroup, we construct a single prompt by filling in these background answers and appending the green-highlighted core question; the same prompt is then used to generate the subgroup-level prediction. 
Concretely, we feed the prompt into the LLM and extract the model’s next-token probabilities over the discrete response options $\{0,\ldots,10\}$, which we treat as the predicted response distribution for that subgroup.

\section{Details for the Generation Datasets}\label{app:wvs_split}

\begin{table}[t]
\centering
\begin{tabular}{ccr}
\toprule
Subset & Usage & Entries \\
\midrule
C1--Q1 & Train & 6,841 \\
C1--Q2 & Val & 1,586 \\
\midrule
C1--Q3 & NewQ & 2,719 \\
C2--Q1 & NewP & 1,179 \\
C2--Q3 & Both & 471 \\
C3--Q1 & - & 1,644 \\
C3--Q3 & - & 660 \\
\bottomrule
\end{tabular}
\caption{WVS split statistics following \citet{TKFT2}. Each entry corresponds to one available (country, question) pair.}
\label{tab:wvs_split_stats}
\end{table}

For direct comparability, we follow the WVS split protocol in \citet{TKFT2}.
We use the first 259 non-demographic WVS questions and partition them into three topical blocks:
Q1 (1--163), Q2 (164--198), and Q3 (199--259).
Countries are divided into three groups:
C2 (Africa): Egypt, Ethiopia, Kenya, Libya, Morocco, Nigeria, Tunisia, Zimbabwe;
C3 (medium-GDP): Malaysia, Thailand, Czechia, Greece, Nigeria, Peru, Colombia, Mexico, Puerto Rico, New Zealand;
and C1 contains the remaining 46 WVS countries.

\noindent\textbf{Train/valid/test.}
Training uses C1--Q1 and validation uses C1--Q2 (same countries as training, unseen questions).
The held-out evaluation in \citet{TKFT2} contains five subsets: C1--Q3, C2--Q1, C2--Q3, C3--Q1, and C3--Q3; in the main text we refer to the first three as (WVS-NewQ, WVS-NewP, WVS-Both), respectively.
\Cref{tab:wvs_split_stats} summarizes the WVS split used in \citet{TKFT2}, reporting the number of available (country, question) pairs in each subset.

\section{Details for the Enhancement Datasets}\label{app:enhancement_details}

\begin{table}[t]
\centering
\begin{tabular}{lrr}
\hline
Dataset & Avg. Respondents & Training Set Size \\ \hline
ESS11   & 38{,}922  & 4{,}000 \\
ESS9    & 45{,}765  & 4{,}000 \\
WVS     & 91{,}854  & 4{,}000 \\
CGSS    & 5{,}044   & 2{,}000 \\
CFPS    & 2{,}435   & 1{,}000 \\ \hline
\end{tabular}
\caption{Enhancement setting statistics. We report the average number of respondents available per question in each dataset and the size of the randomly sampled training set used for enhancement.}
\label{tab:trainingset}
\end{table}

In the \emph{Enhancement} setting, each dataset is used to form multiple \emph{core-question} tasks. 
Concretely, we select 20 core questions per dataset as simulation targets. For each core question, we construct a subgroup-conditioned prompt by pairing it with 5--6 \emph{background questions} that capture respondent characteristics (e.g., demographics and political orientation). 
To choose these background variables in a systematic and scalable manner, we use GPT-5 to screen the full questionnaire and identify 5--6 candidate background questions that are plausible covariates for the target core question (without using any ground-truth answers to the core question). 
We then define subgroups by identical background-answer tuples and use the filled prompt for each subgroup as the input to the LLM (an example is shown in \Cref{sec:appendixprompt}).

To emulate a realistic low-budget survey scenario, we randomly sample a small calibration set from each dataset as training data, while treating the empirical distribution over the full dataset as the ground-truth distribution to be matched. 
\Cref{tab:trainingset} reports, for each dataset, the average number of respondents available per question and the size of the sampled training set used in \Cref{tab:simulation}.

\section{Hyperparameter Settings}

All models are fine-tuned on four NVIDIA RTX 4090 GPUs using the AdamW optimizer with a learning rate of $5\times10^{-5}$. 
We use a batch size of 16 for Qwen3-4B and Llama-8B, and 4 for Qwen-32B. 
To reduce adaptation cost and stabilize calibration on small training sets, we only fine-tune the last Transformer block together with the output softmax layer, while keeping the remaining layers frozen. 
We apply a weight decay of $0.01$ for regularization to mitigate overfitting on the small calibration sets.

\section{Theoretical Analysis}
\label{sec:appendixtheory}

To justify our shift-based estimation method, we formalize an assumption on how different background questions affect the choice distribution.

\paragraph{Assumption 1 (Independent Effects of Background Questions).}
Let $\mathbf{b} = [b_1, b_2, \dots, b_m]$ denote the full background, and let $C$ denote the choice variable.
We assume that the conditional choice distribution factorizes as
\begin{equation}
P(C \mid \mathbf{b})
=
\frac{1}{Z(\mathbf{b})}
\prod_{k=1}^{m} \phi_k(C, b_k),
\label{eq:factorization}
\end{equation}
where each $\phi_k(C, b_k) > 0$ captures the effect of the $k$-th background question on choice $C$, and
\begin{equation}
Z(\mathbf{b})
=
\sum_{c}
\prod_{k=1}^{m} \phi_k(c, b_k)
\label{eq:normalizer}
\end{equation}
is the normalization term ensuring that $P(C \mid \mathbf{b})$ is a valid probability distribution.

This assumption means that the influence of each background question on the choice distribution is separable: each question contributes an independent multiplicative factor, and the full distribution is obtained by combining these factors and renormalizing.

Under Assumption~1, we can make the shift-transfer property precise at the population level.

\paragraph{Theorem 1 (Population-Level Exact Recovery by Shift Transfer).}
Under Assumption~1, fix one background question of interest and denote its value by $x$.
Let $\mathbf{d}$ denote the collection of all remaining background answers.
Then there exist positive functions $\phi(C,x)$ and $\psi(C,\mathbf{d})$ such that
\begin{equation}
P(C \mid x, \mathbf{d})
=
\frac{1}{Z(x,\mathbf{d})}
\phi(C,x)\psi(C,\mathbf{d}),
\label{eq:theorem_factorization}
\end{equation}
where
\begin{equation}
Z(x,\mathbf{d})
=
\sum_c \phi(c,x)\psi(c,\mathbf{d}).
\label{eq:theorem_normalizer}
\end{equation}

Consider two values $x_0, x_1$ of the selected background question, and two settings $\mathbf{d}_1, \mathbf{d}_2$ of the remaining background answers.
Then
\begin{equation}
\begin{aligned}
P(&C \mid x_1, \mathbf{d}_2)= \\
&\operatorname{Normalize}\Bigg(P(C \mid x_0, \mathbf{d}_2)\cdot
\frac{P(C \mid x_1, \mathbf{d}_1)}
     {P(C \mid x_0, \mathbf{d}_1)}
\Bigg).
\end{aligned}
\label{eq:shift_transfer_main}
\end{equation}
Here the normalization is taken over all choices $C$.
Equivalently, the $C$-dependent part of the shift induced by changing $x_0$ to $x_1$ does not depend on the values of the remaining background questions, up to a multiplicative constant.

\begin{proof}
Without loss of generality, suppose the selected background question is the first one, so that
$x = b_1$ and $\mathbf{d} = (b_2, \dots, b_m)$.
Under Assumption~1,
\begin{equation}
P(C \mid x,\mathbf{d})
=
\frac{1}{Z(x,\mathbf{d})}
\phi_1(C,x)\prod_{k=2}^m \phi_k(C,b_k).
\label{eq:proof_start}
\end{equation}
Define
\begin{equation}
\phi(C,x) := \phi_1(C,x),
\qquad
\psi(C,\mathbf{d}) := \prod_{k=2}^m \phi_k(C,b_k).
\label{eq:phi_psi_def}
\end{equation}
Then \eqref{eq:proof_start} becomes
\begin{equation}
P(C \mid x,\mathbf{d})
=
\frac{1}{Z(x,\mathbf{d})}
\phi(C,x)\psi(C,\mathbf{d}),
\label{eq:proof_factorized}
\end{equation}
with
\begin{equation}
Z(x,\mathbf{d})
=
\sum_c \phi(c,x)\psi(c,\mathbf{d}).
\label{eq:proof_factorized_norm}
\end{equation}

Now consider the ratio
\begin{align}
\frac{P(C \mid x_1,\mathbf{d}_1)}{P(C \mid x_0,\mathbf{d}_1)}
&=
\frac{
\frac{1}{Z(x_1,\mathbf{d}_1)}\phi(C,x_1)\psi(C,\mathbf{d}_1)
}{
\frac{1}{Z(x_0,\mathbf{d}_1)}\phi(C,x_0)\psi(C,\mathbf{d}_1)
}
\notag\\
&=
\frac{Z(x_0,\mathbf{d}_1)}{Z(x_1,\mathbf{d}_1)}
\cdot
\frac{\phi(C,x_1)}{\phi(C,x_0)}.
\label{eq:ratio_general}
\end{align}
The $C$-dependent part of this ratio depends only on the change from $x_0$ to $x_1$, and is independent of $\mathbf{d}_1$.

Apply this ratio to $P(C \mid x_0,\mathbf{d}_2)$:
\begin{align}
&P(C \mid x_0,\mathbf{d}_2)
\cdot
\frac{P(C \mid x_1,\mathbf{d}_1)}
     {P(C \mid x_0,\mathbf{d}_1)}
\notag\\
&=
\left(
\frac{1}{Z(x_0,\mathbf{d}_2)}
\phi(C,x_0)\psi(C,\mathbf{d}_2)
\right)
\notag\\
&\qquad\cdot
\left(
\frac{Z(x_0,\mathbf{d}_1)}{Z(x_1,\mathbf{d}_1)}
\cdot
\frac{\phi(C,x_1)}{\phi(C,x_0)}
\right)
\notag\\
&=
\frac{Z(x_0,\mathbf{d}_1)}
     {Z(x_1,\mathbf{d}_1)\,Z(x_0,\mathbf{d}_2)}
\phi(C,x_1)\psi(C,\mathbf{d}_2).
\label{eq:shift_application_general}
\end{align}

The right-hand side of \eqref{eq:shift_application_general} is proportional to
\begin{equation}
P(C \mid x_1,\mathbf{d}_2)
=
\frac{1}{Z(x_1,\mathbf{d}_2)}
\phi(C,x_1)\psi(C,\mathbf{d}_2).
\label{eq:target_distribution}
\end{equation}
Therefore,
\begin{equation}
P(C \mid x_1,\mathbf{d}_2)
\propto
P(C \mid x_0,\mathbf{d}_2)
\cdot
\frac{P(C \mid x_1,\mathbf{d}_1)}
     {P(C \mid x_0,\mathbf{d}_1)}.
\label{eq:proportional_general}
\end{equation}
After normalization over all choices $C$, we obtain
\begin{equation}
\begin{aligned}
P(&C \mid x_1,\mathbf{d}_2)=\\
&\operatorname{Normalize}\Bigg(
P(C \mid x_0,\mathbf{d}_2)\cdot
\frac{P(C \mid x_1,\mathbf{d}_1)}
     {P(C \mid x_0,\mathbf{d}_1)}
\Bigg).
\end{aligned}
\label{eq:normalized_general}
\end{equation}
This proves the claim.
\end{proof}

Theorem~1 shows that, under the factorization assumption, the effect of changing one background answer can be transferred across different settings of the remaining background questions at the population level.
That is, if the true conditional distributions are known exactly, then the target distribution can be recovered exactly by applying the corresponding shift and renormalizing.

In finite-sample settings, however, one does not observe the population distributions directly and must instead estimate them from data.
Therefore, Theorem~1 should be understood as an identification result rather than a finite-sample guarantee.
In practice, pooling information across related backgrounds may improve statistical efficiency, but such gains depend on the estimator and sample size, and are not implied by the theorem itself.

\end{document}